\documentclass[letterpaper, 10 pt, conference]{ieeeconf}
\usepackage{cite}
\usepackage[utf8]{inputenc}

\usepackage{hyperref}
\usepackage{caption}

\usepackage{subcaption} 
\usepackage{graphicx}
\usepackage{epstopdf}
\graphicspath{{../pdf/}{../jpeg/}{../img/}}
\DeclareGraphicsExtensions{.pdf,.jpeg,.png,.eps}

\usepackage{listings}
\usepackage{verbatim}
\usepackage[cmex10]{amsmath}
\usepackage{array}
\usepackage{url}

\usepackage{xcolor}%


\begin{document}
\title{\vspace{-0.86cm}\LARGE \bf Towards a DSL for Perception-Based Safety Systems\vspace{-0.25cm}}
\author{Johann Thor Mogensen Ingibergsson$^{*}$ and Stefan-Daniel Suvei$^{*}$ and\\ Mikkel Kragh Hansen$^{\dag}$ and Peter Christiansen$^{\dag}$ and Ulrik Pagh Schultz$^{*}$ \\
$^{*}$ University of Southern Denmark, Campusvej 55, 5230 Odense M, Denmark\\
$^{\dag}$ Department of Engineering, Aarhus University, Finlandsgade 22, 8200 Aarhus N, Denmark\\
Email: \{jomo$|$stdasu$|$ups\}@mmmi.sdu.dk or \{pech$|$mkha\}@eng.au.dk}

\maketitle


\section{Introduction}
\label{sec:Introduction}
Agriculture sees a high number of injuries and
fatalities, even in developed countries~\cite{agricultural_statistics_board_agricultural_2013}. Field robots are a solution to this problem, but the issue with field robots is that they are prone to failure~\cite{reichardt_software_2013}, which has led to the use of Model-Driven Engineering~(MDE). In robotics, MDE is used to improve development time and reliability, for example in regards to behavior~\cite{adam_towards_2014}. 
In order to ensure safe autonomous operation, robust and reliable risk detection and obstacle avoidance must be performed.
In this regard, field robots are highly dependent on perception sensors and algorithms to understand and react on the environment.
The robot has to observe a large area; it must be fast, reliable and robust; it is constrained to function with low computational resources due to embedded hardware; and it might have lower priority than control and must not jam~\cite{ingibergsson_dslrob_2015}. This imposes severe constraints on the software that interacts with sensors. 
Therefore, we propose a DSL~\cite{ingibergsson_dslrob_2015} for safety concerns in perception systems. 
Our current research question is to investigate if the safety rules should be split between distinct domains (behavior~\cite{adam_towards_2014} and perception~\cite{ingibergsson_dslrob_2015}), or combined into one (this paper).


\section{Method and Experimental Setup}
\label{sec:MethodandExperimental}

\begin{figure}[thpb]
	\centering
	\fbox{\includegraphics[width=0.236\textwidth]{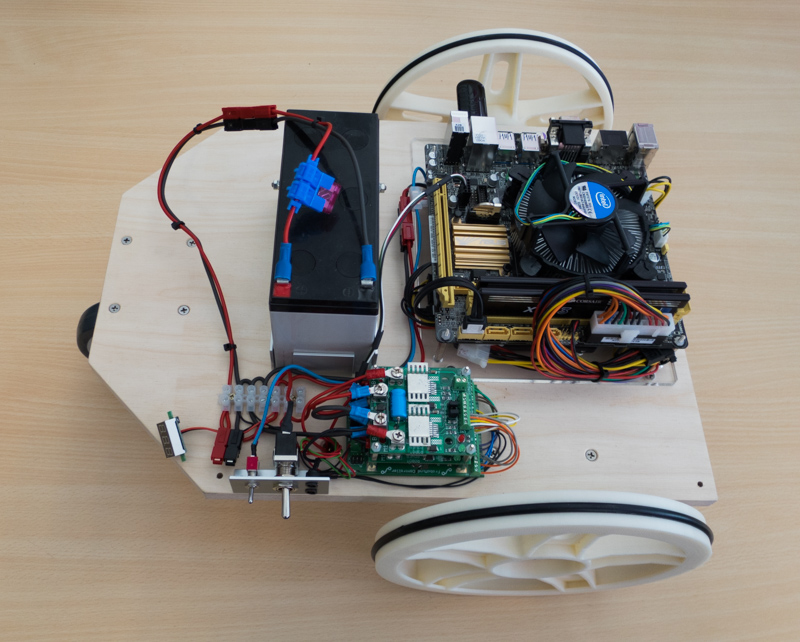}}
  \fbox{\includegraphics[width=0.2147\textwidth]{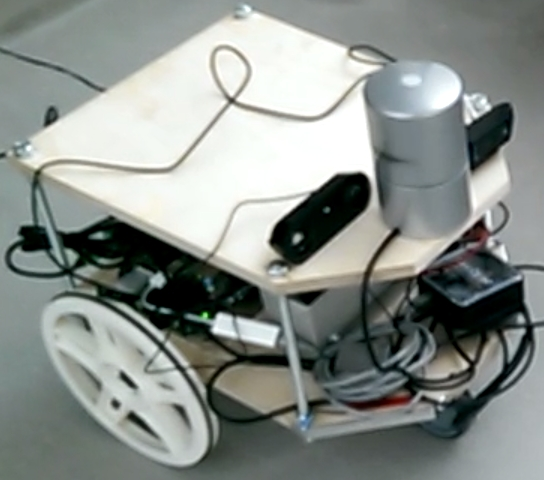}}
	\caption{Field robot prototype with lidar and camera sensors.}
	\label{fig:fieldrobotlabel}
\end{figure}

For investigating the research question, we built a field robot prototype (Fig.~\ref{fig:fieldrobotlabel}) to experiment with language design. First step is a test case implemented without MDE. To this end, we have extended a simple robot called Frobit, shown above, with non-trivial safety based on different sensors to prototype an agricultural field robot. The sensors are a Velodyne HDL-32E lidar with 32 channels for 360 degree 3D view and a Logitech C920 HD webcam.
The prototype is designed for operation within an orchard. The lidar is used for detecting the orchard row to enable general navigation, and for object detection in front of the robot to enable obstacle avoidance~\cite{kragh_mikkel_hansen_2015}. The distances from the obstacles to the robot are used to adjust to the speed. The rows are found using a Hough transform to enable the robot to position itself in the middle of the row. 
The camera detects humans using a C++ implementation of a pedestrian detector by Dollár ~\cite{pdollar}. A human position is estimated using the tilt of the camera, the bounding box position and a flat surface assumption. 
If it detects a human, the robot will sound warnings and limit the maximum speed according to the distance to the human, and ultimately bring the robot to a full stop.

\begin{figure}[thpb]
	\centering
  \fbox{\includegraphics[width=0.45\textwidth]{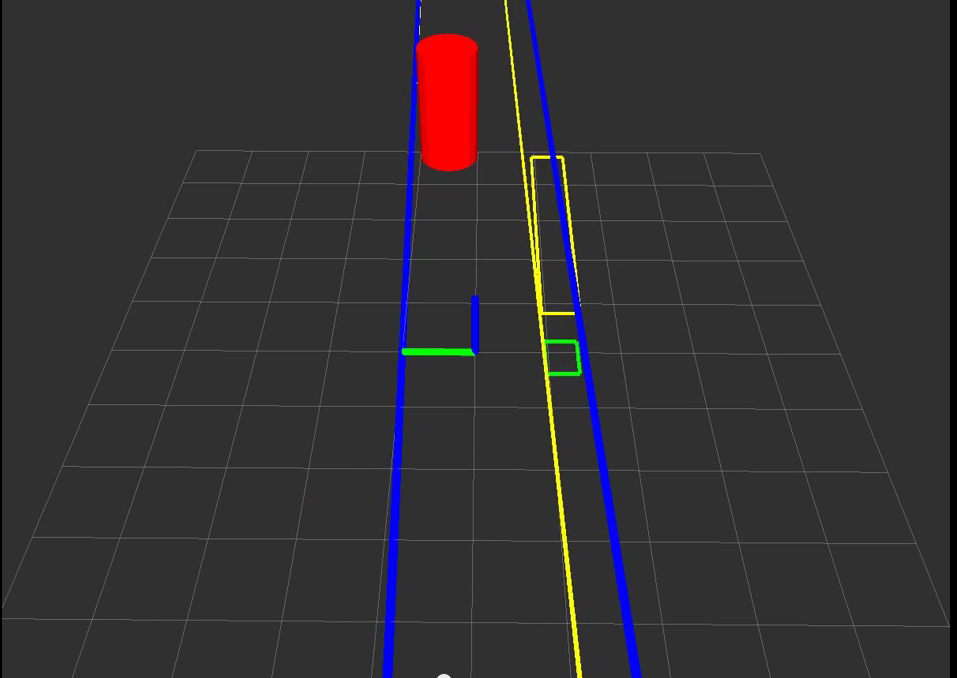}}
	\caption{Example of row navigation and obstacle avoidance. The robot has detected the rows (blue lines), a close obstacle (yellow rectangle), and a pedestrian (red cylinder). It adjusts its navigation (yellow line) and decreases its speed accordingly.}
	\label{fig:demonstration}
\end{figure}

Fig.~\ref{fig:demonstration} visualizes an example from an actual field trial. The red cylinder denotes a detected pedestrian, the blue lines denote the detected rows, and the yellow rectangle denotes the closest obstacle detected by the lidar.
The traversable width of the row is decreased by the algorithm (indicated by the yellow line), such that the robot will effectively navigate around the closest obstacle.
Due to a small distance to the detected pedestrian, the robot sounds a warning and decreases its speed.

\section{Language Design}
\label{sec:LanguageDesign}
We propose ideas of how to achieve sufficient safety levels
for the perception system and the field robot as a whole. We want to create a DSL that combines behavioral safety~\cite{adam_towards_2014} with our proposed perception safety system~\cite{ingibergsson_dslrob_2015}. An outline of the DSL utilizing safety from both systems, can be seen below: \\
\\

\begin{lstlisting}
exists p in camera.all(pedestrian):
    distance(p) < 1m { sound emergency; stop; }
    distance(p) < 3m { sound move_away; cap_speed; }
    distance(p) < 5m { sound please_move_away; }
hist h = histogram(camera.image, bins = 10, normalized=true):
    size(x in h.bins: size(x)>0)/size(h.bins)<0.3 { cap_speed; }
    size(x in h.bins: size(x)>0)/size(h.bins)<0.1 { stop; }
    max(x in h.bins: size(x) > 0) - min(x in h.bins: size(x) > 0) < 1000px { stop; }
exists o in laser.all(Objects):
    distance(o) < 5m { cap_speed; }
    distance(o) < 0.5m { sound emergency; stop; }
lasers a in lasers(alive):
   count(a) < 32 { cap_speed; }
   count(a) < 26 { stop; }
\end{lstlisting}

Here, the output of the different sensors can be subjected to rules, that will enable the system to analyze the trustworthiness of the sensor and act accordingly. The first rule uses a pedestrian detector to calculate the distance to the nearest human and will then use cognitive safety to react, i.e., try to handle the situation before the emergency system kicks in. 
In addition to this, a functional layer performs a histogram analysis of the incoming images, verifying that the camera sensor is working.
Similarly for a lidar, cognitive safety and functional safety are divided between different functionalities. 
In the first rule, the lidar detects objects using information from all available laser beams.
Depending on the distance of an object, the speed of the robot is decreased accordingly.
For the functional layer, it is possible to look at graceful degradation for the lidar, as it consists of many laser beams that provide a certain degree of redundancy. This is indicated by ``\textit{lasers(alive)}''.
Functionality corresponding to the above code has been tested on the robot. The code {\em developed without the proposed language}, is 8454 lines of C++ and Python ROS code. The language that we propose could, if implemented, cut down the code to 14 lines. The next steps will be to follow MISRA~\cite{misra_misra-c_2012} rules to extend the trustworthiness of the code into the domain of standards.

\bibliographystyle{ieeetr}

\end{document}